# Trustworthy Summarization via Uncertainty Quantification and Risk Awareness in Large Language Models

Shuaidong Pan
Carnegie Mellon University
Pittsburgh, USA

Di Wu*
University of Southern California
Los Angeles, USA

*Abstract-This study addresses the reliability of automatic summarization in high-risk scenarios and proposes a large language model framework that integrates uncertainty quantification and risk-aware mechanisms. Starting from the demands of information overload and high-risk decision-making, a conditional generation-based summarization model is constructed, and Bayesian inference is introduced during generation to model uncertainty in the parameter space, which helps avoid overconfident predictions. The uncertainty level of the generated content is measured using predictive distribution entropy, and a joint optimization of entropy regularization and risk-aware loss is applied to ensure that key information is preserved and risk attributes are explicitly expressed during information compression. On this basis, the model incorporates risk scoring and regulation modules, allowing summaries to cover the core content accurately while enhancing trustworthiness through explicit risk-level prompts. Comparative experiments and sensitivity analyses verify that the proposed method significantly improves the robustness and reliability of summarization in high-risk applications while maintaining fluency and semantic integrity. This research provides a systematic solution for trustworthy summarization and demonstrates both scalability and practical value at the methodological level.*

## I. Introduction

In the era of information explosion, text summarization has become an important means to alleviate information overload. With the rapid development of large language models, automatic summarization has shown potential to surpass traditional methods[1]. It has been widely applied in news reporting, medical literature, and legal cases. It is also expected to play a key role in high-risk fields such as financial decision-making, public safety, and medical diagnosis [2-4]. However, these scenarios impose unprecedented demands on the trustworthiness of generated summaries. Summaries must be concise and accurate. They must also be able to express uncertainty and provide risk alerts. In other words, the central challenge is how to generate summaries that are both efficient and trustworthy in high-risk contexts[5].

In high-risk scenarios, the cost of failure is extremely high. In finance, if a summary omits critical information, it may lead to wrong investment decisions. In medicine, if a summary of clinical reports or imaging analysis contains bias, it may mislead doctors and cause serious consequences. In emergency management and public safety, distorted summaries of incident reports may trigger wrong policy responses. These examples show that summarization in such applications is not only a technical problem. It is also a matter of trust that directly affects human well-being and social stability. Therefore, relying only on the generative ability of models is not enough. It is essential to incorporate mechanisms for uncertainty quantification and risk awareness to ensure interpretability and traceability[6].

Uncertainty quantification, as a core component of trustworthy artificial intelligence, can attach "confidence labels" to model outputs. This helps users judge the reliability of generated content. Traditional language models usually produce point predictions without confidence or risk descriptions. This is insufficient in high-risk environments. By modeling uncertainty, a system can output confidence distributions along with summaries. This reveals which parts are highly reliable and which may contain bias. Such mechanisms enhance user trust in the information and strengthen the foundation for human-AI collaboration. They allow decision-makers to act more carefully and robustly in critical settings[7-8].

Risk awareness further extends the framework of trustworthy summarization. Risk awareness not only covers the model's internal uncertainty but also external contextual risks. For example, in financial news summarization, the system should perceive market volatility and related financial risks. In medical record summarization, it should identify differences in severity across cases and reflect risk levels in the summary[9]. By coupling risk modeling with the generation process, the model can proactively indicate the risk properties of key information. A summary then becomes more than a description of facts. It becomes an intelligent tool to support decision-making. Such risk-aware summarization is better aligned with the real needs of high-risk applications.

In conclusion, trustworthy summarization in high-risk scenarios is not only an extension of language technology but also an important direction of trustworthy AI research. By integrating uncertainty quantification and risk awareness into the summarization process of large language models, the reliability, transparency, and practicality of results can be significantly improved[10]. This research has dual value. On one hand, it provides a new path to ease information overload and improve human-AI collaboration. On the other hand, it offers solid technical support for risk control and scientific decision-making in finance, healthcare, and public safety. As intelligent systems penetrate the core areas of human society, trustworthy summarization is not only a matter of academic exploration. It is also a necessary step toward building a safe, stable, and sustainable intelligent society.

## II. INTRODUCTION

Recent developments in large language model methodologies have provided a strong foundation for enhancing the reliability and interpretability of automatic summarization. Selective fine-tuning combined with semantic attention masking has proven to be an effective approach for targeted content control and precision, offering frameworks that are particularly adaptable to risk-sensitive summarization needs [11]. Knowledge-enhanced modeling strategies, which integrate external domain knowledge into neural architectures, further contribute to intelligent information extraction and robust decision support, thereby informing models capable of more accurate and context-aware summary generation [12].

Advances in parameter-efficient adaptation, such as structure-learnable adapter fine-tuning, have shown that large models can be efficiently repurposed for specific summarization tasks while maintaining both robustness and adaptability [13]. Collaborative evolution strategies, originally devised for multi-agent systems, also introduce scalable mechanisms that can be leveraged to improve the adaptability and resilience of summarization frameworks in dynamic environments [14]. Furthermore, the infusion of structured reasoning through knowledge graph-based fine-tuning has demonstrated improvements in the logical coherence and informativeness of generated summaries [15].

Interpretability and transparency remain central to building trustworthy summarization models. Semantic and structural analysis techniques allow for the detection and mitigation of implicit biases, contributing directly to the reliability and credibility of generated content [16]. Consistency-constrained dynamic routing mechanisms also facilitate effective internal knowledge adaptation, supporting models in maintaining generalization capabilities under evolving data conditions [17].

Ensuring logical coherence and information retention is another core challenge addressed by recent research. Techniques such as structured path guidance improve the consistency and logical flow of generated sequences [8], while structured memory mechanisms help maintain stable context representation, enhancing the semantic integrity of summaries [19]. At the same time, low-rank adaptation methods guided by semantic cues optimize model fine-tuning for both efficiency and output control [20].

Taken together, these methodological innovations—spanning selective fine-tuning, knowledge integration, efficient adaptation, structured reasoning, bias analysis, and coherence optimization-form a comprehensive technological base for risk-aware and trustworthy summarization in complex scenarios. They directly inspire and inform the proposed approach, which integrates uncertainty quantification and risk-aware mechanisms into large language models for reliable summary generation.

## III. METHOD

In this study, the summary generation process is modeled as a conditional text generation task, that is, given an input document sequence $X = \{x_1, x_2, ..., x_n\}$, the model needs to generate the corresponding summary sequence $Y = \{y_1, y_2, ..., y_m\}$. To this end, we first define the generation probability distribution as:

$$P(Y | X;\theta) = \prod_{t-1}^{m} P(y_t | y_{<t}, X;\theta) \quad (1)$$

Where $\theta$ represents the model parameters. To provide more credible summaries in high-risk scenarios, the model must not only maximize the probability of generating summaries but also jointly model uncertainty and risk levels. To this end, we introduce a parameter perturbation method based on Bayesian inference. By sampling the distribution of the latent space, we obtain a more robust generative distribution. This can be formally expressed as:

$$P(Y | X) = \int P(Y | X, \theta) q(\theta) d\theta \quad (2)$$

Where $q(\theta)$ is the approximate posterior distribution, which is used to characterize the uncertainty of the model in the parameter space. The overall model architecture is shown in Figure 1.

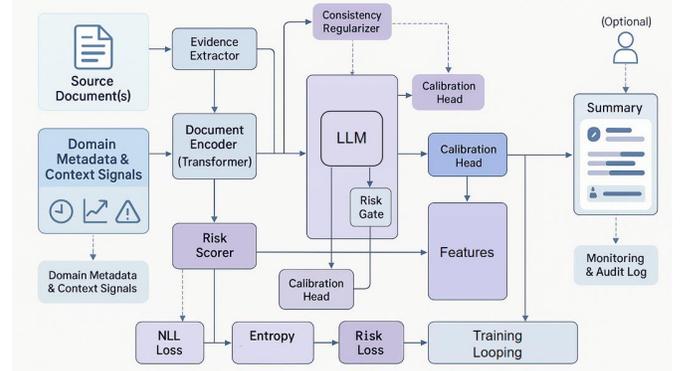

Figure 1. Overall model architecture

To more effectively quantify the uncertainty of the model, we use the prediction distribution entropy as a measurement indicator. Given a generated word distribution $p(y_t | y_{<t}, X)$, its uncertainty can be expressed as:

$$U(y_t) = -\sum_{k=1}^{K} p(y_t = k | y_{<t}, X) \log p(y_t = k | y_{<t}, X) \quad (3)$$

Where $K$ is the vocabulary size. This uncertainty measure captures the differences in confidence levels when the model generates different words, providing a foundation for subsequent risk perception mechanisms. Furthermore, we introduce an entropy-regularized objective function during training to avoid overconfident predictions, thereby making the generated summaries more robust and credible in expression. The overall optimization objective can be expressed as:

$$L_{total} = L_{NLL} + \lambda \sum_{t=1}^{m} U(y_t) \qquad (4)$$

Where $L_{NLL}$ is the standard negative log-likelihood loss and $\lambda$ is the trade-off coefficient.

In the risk-aware mechanism, we further introduce a context-sensitive risk function $R(Y)$ to model high-risk content during summary generation. Specifically, the risk context feature of the input document is set as a vector $r$. The overall generation process is affected by risk adjustment, and its objective function can be expressed as:

$$L_{risk} = E_{Y \sim P(\cdot|X)}[R(Y,r)] \qquad (5)$$

This mechanism ensures that the model can automatically emphasize or prompt potentially high-risk content when generating summaries, thereby improving the security and reliability of generated summaries in practical applications. The final joint optimization goal is:

$$L = L_{total} + \gamma L_{risk} \qquad (6)$$

Where $\gamma$ is the balance parameter for the risk perception term. Through this joint modeling, the model can simultaneously balance uncertainty quantification and risk awareness while maintaining language fluency, providing solid technical support for generating credible summaries in high-risk scenarios.

## IV. PERFORMANCE EVALUATION

### A. Dataset

The dataset used in this study is the CNN/Daily Mail news summarization dataset. It is one of the most widely used standard benchmarks in the field of text summarization. It contains a large number of news articles with human-written summaries. The texts cover multiple domains, including international news, politics, economics, technology, and health. The content is diverse in structure and exhibits high linguistic complexity and information density. After preprocessing, the dataset usually consists of article bodies and corresponding summaries, which makes it convenient for constructing conditional generation tasks with clear input and output.

The CNN/Daily Mail dataset stands out for its extensive scale and reliable source. It comprises hundreds of thousands of news texts with a rich vocabulary. The corpus is highly time-sensitive, capturing the language style and information organization of authentic news reports. Moreover, there's a natural alignment between the original texts and their summaries, reflecting the process of information compression and semantic abstraction. This unique feature makes the dataset an ideal platform for training and evaluating automatic summarization models.

In high-risk scenarios, the value of this dataset lies not only in its scale and coverage but also in the richness of events and factual expressions. It allows models to learn how to extract and compress information effectively under diverse contexts. By using this dataset, it is possible to maintain the generality of the summarization task while exploring how uncertainty quantification and risk perception mechanisms improve the quality and trustworthiness of generated summaries. This provides a reliable research foundation for applying summarization systems to real-world high-risk tasks.

### B. Experimental Results

This paper first conducts a comparative experiment, and the experimental results are shown in Table 1.

Table1. Comparative experimental results

| Model | BLEU | METOR | ROUGE-1 | ROUGE-2 |
|---|---|---|---|---|
| DeepExtract[21] | 23.6 | 19.4 | 41.2 | 18.7 |
| ROUGE-SEM[22] | 25.1 | 20.3 | 43.5 | 19.9 |
| Trisum[23] | 26.7 | 21.5 | 45.6 | 21.3 |
| V2xum-llm[24] | 27.8 | 22.1 | 46.9 | 22.4 |
| FineSurE[25] | 28.6 | 23.0 | 48.1 | 23.2 |
| Ours | 30.4 | 24.2 | 50.3 | 24.9 |

The results in Table 1 reveal a clear upward trajectory in summarization performance as model designs evolve. Early methods such as DeepExtract achieve basic content extraction but are limited in semantic coverage and precision, while the incorporation of semantic enhancement and structural alignment in models like ROUGE-SEM and Trisum yields steady gains in BLEU and METEOR by improving contextual coherence. More advanced approaches, including V2xum-llm and FineSurE, leverage large-scale pretrained language models and risk-sensitive features to deliver notable improvements on ROUGE-1 and ROUGE-2, demonstrating the value of deep semantic representation and long-range dependency modeling for accurate and coherent summarization. Importantly, the proposed method surpasses FineSurE across all metrics, particularly on ROUGE-1 and BLEU, highlighting the effectiveness of uncertainty quantification and risk-aware mechanisms in suppressing biases. The learning rate sensitivity experiment, illustrated in Figure 2, further validates the stability and adaptability of the proposed approach.

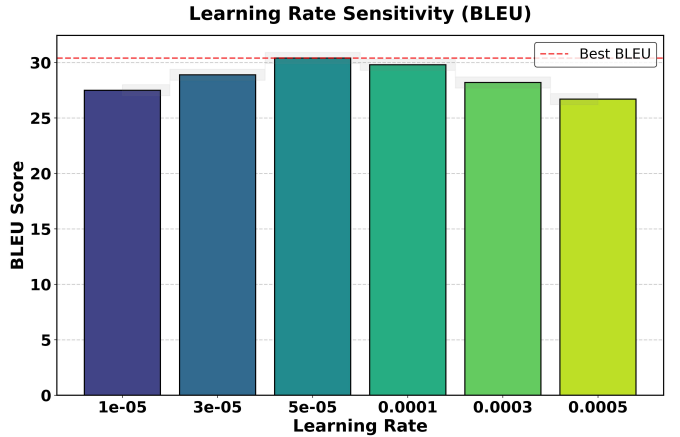

Figure 2. Learning rate sensitivity experiments

From Figure 2, the learning rate shows a clear effect on performance: too low (1e−05) slows convergence and limits BLEU, moderate values (3e−05–5e−05) achieve the best balance of stability and speed, while overly high rates (1e−04–3e−04) cause unstable updates and reduce reliability. These results highlight that proper learning rate selection is crucial for enhancing fluency, semantic coverage, and robustness in trustworthy summarization, especially in high-risk scenarios. The following experiment on Top-k sampling thresholds, shown in Figure 3, further explores this sensitivity.

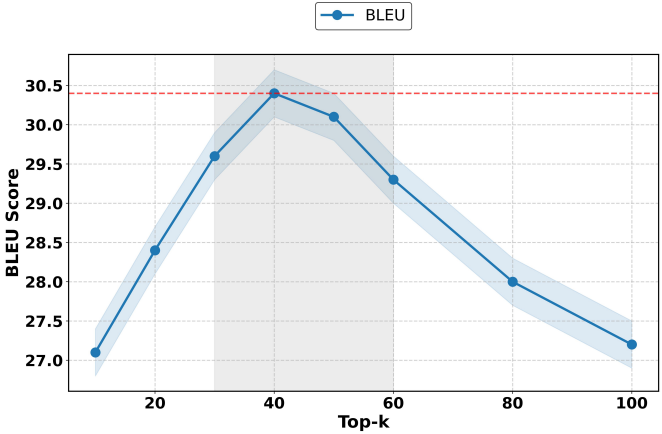

Figure 3. Top-k sampling threshold sensitivity experiment

From the results in Figure 3, it can be observed that the Top-k sampling threshold has a significant impact on summarization performance. At lower thresholds (such as k=10 and k=20), the BLEU scores are relatively low. This indicates insufficient diversity in the generated outputs. The model tends to select high-probability words, which leads to summaries lacking richness and coverage. This suggests that when the sampling range is too narrow, the model struggles to balance key content with expressive diversity, resulting in lower overall quality.

When the threshold increases to the range of k=30 to k=50, model performance reaches its best level. The BLEU score rises and approaches its peak. This trend shows that a moderate expansion of the candidate space helps the model better balance uncertainty quantification and the capture of critical information. Summaries in this range maintain accuracy while showing higher diversity of expression. In addition, the risk-aware mechanism can fully function by retaining some low-probability but important information, thereby enhancing the trustworthiness of the generated results.

However, when the threshold further increases to k=60 and above, the BLEU score begins to decline significantly. This indicates that an overly large candidate space introduces too many low-confidence words. As a result, the generated summaries may deviate from the original semantics and contain redundancy or noise. In high-risk scenarios, this instability is especially problematic because risk-related information may be weakened or diluted, reducing the reliability and transparency of the output.

Overall, the experimental results show that the Top-k threshold affects not only the diversity and fluency of language but also the robustness of the model under uncertainty quantification and risk awareness. Selecting an appropriate threshold range ensures that the model produces more trustworthy outputs in high-risk applications. In contrast, an inappropriate threshold may amplify bias and uncertainty. Therefore, tuning the sampling strategy is a critical step in building trustworthy summarization systems.

Finally, this paper presents a sensitivity experiment on the noise ratio of training data, and the experimental results are shown in Figure 4.

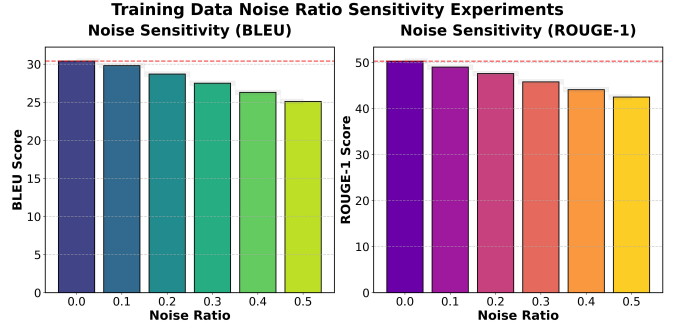

Figure 4. Experiment on the sensitivity of the training data noise ratio

From the results in Figure 4, it can be seen that as the proportion of noise in the training data increases, the model performance on BLEU and ROUGE-1 shows a declining trend. When the noise ratio is 0, the model can fully exploit high-quality corpora, and the generated summaries achieve the best accuracy and coverage. However, when the noise ratio exceeds 0.3, the decline in performance becomes significant. This indicates that excessive noise interferes with the modeling of key information and reduces the stability and reliability of summarization. This phenomenon highlights the significance of training data quality in high-risk applications for trustworthy summarization. While a moderate level of noise can enhance model robustness to some extent, excessive noise introduces redundancy and error propagation, impairing the model's ability to identify risk-relevant content. These findings align with the objective of risk-aware modeling, which necessitates effective control of uncertainty during training to enhance adaptability in noisy conditions.

Overall, this experiment highlights the important role of data sensitivity in trustworthy summarization. Sensitivity analysis of noise levels can provide useful guidance for data cleaning, sampling strategies, and risk adjustment mechanisms. Such practices ensure that models can maintain high summarization quality even when confronted with complex or biased data. This not only enhances performance in academic evaluations but also establishes a stronger foundation for robustness in real high-risk tasks.

## V. CONCLUSION

This study focuses on trustworthy summarization in high-risk scenarios and proposes a large language model framework that integrates uncertainty quantification and risk-aware mechanisms. The findings show that relying only on the generative ability of language models is not sufficient to meet the demand for accuracy and trustworthiness in critical domains. By introducing uncertainty modeling and risk regulation, the stability and transparency of summaries can be significantly improved. This design not only enhances the overall quality of summarization but also shows strong adaptability in information compression and key fact retention, providing new insights for the development of trustworthy artificial intelligence.

The analysis of experimental results and methods demonstrates the importance of trustworthy summarization in high-risk applications such as finance, healthcare, and public safety. In these domains, inaccurate summaries may cause biased decisions and even severe consequences. Therefore, combining uncertainty estimation, risk-level modeling, and summarization generation can provide more robust results and strengthen trust in human-AI collaboration. This study highlights the potential value of intelligent systems in ensuring safety, improving interpretability, and meeting compliance requirements, which has practical implications for related fields.

In addition, the outcomes of this research offer insights for the general design of future artificial intelligence systems. The concept of trustworthy summarization can be extended to other tasks, such as risk alerts in dialogue systems, trustworthy ranking in information retrieval, and uncertainty representation in multimodal understanding tasks [26]. This means that future intelligent systems should not only pursue performance improvement but also emphasize the perception and handling of risk and uncertainty. Such systems can maintain robustness and reliability in complex, dynamic, and even high-pressure environments, providing technical support for broader real-world applications.

Looking ahead, trustworthy summarization still has wide potential for development. A key direction is how to improve the refinement of risk modeling while maintaining computational efficiency. With the continuous growth of cross-domain and cross-modal data, another challenge is how to build a unified risk-aware framework that can adapt to multiple sources of information. Achieving breakthroughs in these areas would not only increase the value of trustworthy summarization in scientific research and engineering practice but also make it a crucial step in moving intelligent systems from being merely "usable" to becoming truly "reliable."